\documentclass[review,sort&compress]{elsarticle}
\usepackage{amssymb}
\usepackage{amsmath,amsfonts}
\usepackage[caption=false,font=normalsize,labelfont=sf,textfont=sf]{subfig}
\usepackage{algorithmic}
\usepackage{algorithm}
\usepackage{array}
\usepackage{verbatim}
\usepackage{graphicx}
\usepackage{ifpdf}
\usepackage{booktabs}
\usepackage{color} 
\usepackage{multirow}

\journal{Knowledge-Based Systems}

\begin{document}

\begin{frontmatter}

\title{Intention Action Anticipation Model with Guide-Feedback Loop Mechanism}

\author[label1]{Zongnan~Ma}
\author{Fuchun~Zhang\corref{cor1}\fnref{label1}}
\ead{yadxzfc@yau.edu.cn}
\cortext[cor1]{Corresponding author}
\author{Zhixiong~Nan\corref{cor1}\fnref{label2}}
\ead{nanzx@cqu.edu.cn}
\author[label3]{Yao~Ge}

\affiliation[label1]{organization={School of Physics and Electronic, Yan'an University},
            city={Yan'an},
            postcode={716000},
            country={China}}

\affiliation[label2]{organization={College of Computer Science, Chongqing University},
            city={Chongqing},
            postcode={400044},
            country={China}}

\affiliation[label3]{organization={Continental-NTU Corporate Laboratory, Nanyang Technological University},
            postcode={637553},
            country={Singapore}}
            
\begin{abstract}
Anticipating human intention from videos has broad applications, such as automatic driving, robot assistive technology, and virtual reality.
This study addresses the problem of intention action anticipation using egocentric video sequences to estimate actions that indicate human intention.
We propose a Hierarchical Complete-Recent (HCR) information fusion model that makes full use of the features of the entire video sequence (i.e., complete features) and the features of the video tail sequence (i.e., recent features). 
The HCR model has two primary mechanisms. The Guide-Feedback Loop (GFL) mechanism is proposed to model the relation between one recent feature and one complete feature. Based on GFL, the MultiComplete-Recent Feature Aggregation (MCRFA) module is proposed to model the relation of one recent feature with multiscale complete features. Based on GFL and MCRFA, the HCR model can hierarchically explore the rich interrelationships between multiscale complete features and multiscale recent features. Through comparative and ablation experiments, we validate the effectiveness of our model on two well-known public datasets: EPIC-Kitchens and EGTEA Gaze+.
\end{abstract}


\end{frontmatter}

\section{Introduction}
\begin{figure}[!t]
	\centering
	\includegraphics[width=0.7\textwidth]{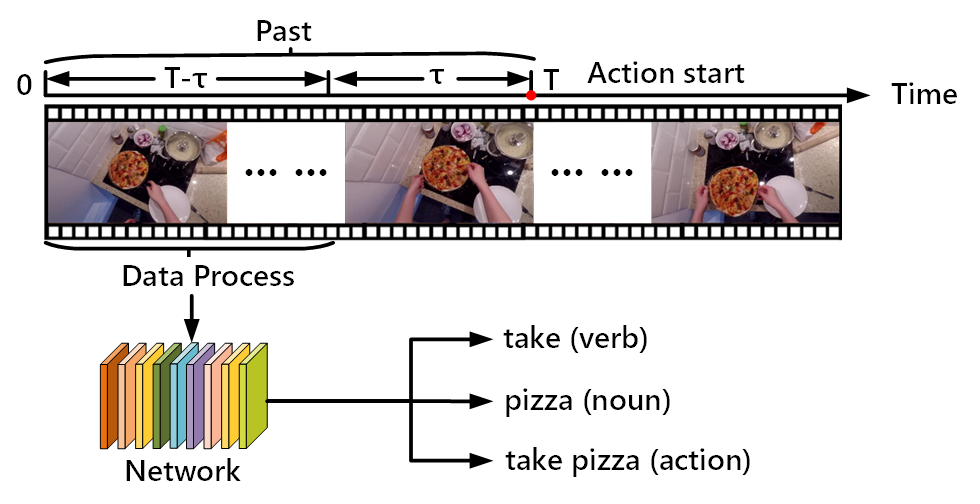}
   \vspace{-5mm}
	\caption{Visualization of intention action anticipation. Anticipation time $\tau$ is how much in advance the intention action has to be anticipated.}
  \vspace{-3mm}
	\label{fig_start}
\end{figure}

Human intention anticipation has wide applications in diverse fields, such as assistive robots \cite{CRAMER2018255, Schydlo2018AnticipationIH}, autonomous driving \cite{Bae2021DisentangledMG, Fang2020TPNetTP, degeest2016online}, patient healthcare \cite{Gershov2024TowardsAA, s22030831}, and surveillance systems \cite{9025515}. Consequently, an increasing number of researchers are focusing on this problem. This study addresses the problem of human intention action anticipation using egocentric videos collected by humans performing daily activities. As shown in Fig.~\ref{fig_start}, the model uses the input video sequence before the anticipation time (i.e., $[0, T-\tau]$) to obtain the anticipated results in the future time (i.e., $[T, ]$). 

Most existing methods \cite{9726229, Miech2019, furnari2019, Girdhar2021A,  2021TransAction, 2021SlowFast, 9352754,9003286,sener2020, Nawhal2022R} have modeled both spatial features and temporal contexts, as both aspects are essential to accurately anticipate human intention. To deeply extract the spatial information, these researchers emphasized considering the various characteristics of data in different manners should be considered, such as RGB modality expressing the entire scene, object modality (OBJ) detailing crucial zones in the scene, and optical flow (FLOW) modality signaling human motion. 
Some methods \cite{9726229, 9003286} use human skeletal modality to acquire rich motion information.
Through the fusion of the above multimodal information, comprehensive spatial information can be obtained.
However, temporal contexts have not received as much attention as spatial information.
Most researchers \cite{furnari2019, Girdhar2021A,   2021TransAction, 2021SlowFast} only adopted a straightforward mechanism to process temporal contexts, which cannot fully use temporal information. 
Few researchers \cite{Miech2019, sener2020, Nawhal2022R} have realized the importance of temporal information and they extracted long-term features and short-term features from video clips. 
However, they have treated these features simply without adequately leveraging their strengths of the features. 
This oversight causes their approach to fail to fully exploit the unique information of different time scales. 

Based on the aforementioned observation, the existing methods have some shortcomings in exploiting temporal information. 
\textbf{\#1)} Human activities are complex, dynamic, and random, sometimes releasing misleading signals for intention anticipation. For example, when a human’s hand is near an object, it is unclear whether the human will approach or pull away from the object.
\textbf{\#2)} The time duration before a human performs the intended action is inconsistent. In some cases, a 1-s duration has revealed human intention. In other cases, understanding human intention might require a 10-s duration.
In particular, we analyze existing methods that treat all video frames with equal importance weights. However, in most cases, the frames at the beginning and end of a video are not equally important. This oversight can make it challenging to address challenge \textbf{\#1}. In addition, existing methods model temporal contexts at a fixed scale, making it challenging to address complex human activities occurring over different periods. This oversight may lead to difficulty in addressing challenge \textbf{\#2}.

We address these deficiencies and overcome these challenges by proposing novel approaches that emphasize the importance of varying temporal scales. Our strategy involves modeling both multiscale and global/local temporal contexts. A multiscale temporal context allows for considering variations in the duration required to comprehend intentions. Furthermore, a more robust and nuanced analysis can be achieved by considering the wider context of the entire video sequence (global temporal context) and the temporal relationships within smaller segments (local temporal context).

Based on the above analysis, we propose a Hierarchical Complete-Recent (HCR) information fusion model. \textbf{First}, we argue that frames in a video have different importance weights, and those at the video tail are more essential for human intention prediction in most situations. Thus, we extract \textit{complete features} from the entire video and \textit{recent features} from the frames at the tail of the video. \textbf{Second}, we believe that a multitemporal scale encoder can extract more powerful temporal contexts and is necessary for human intention prediction in open scenes, thus both \textit{complete features} and \textit{recent features} are a set of features extracted in multi-scale temporal dimensions. \textbf{Third}, we propose that a hierarchical-structured model could well organize different levels of features. Thus, we propose a guide-feedback loop (GFL) module to model the one recent feature-one complete feature relation and a MultiComplete-Recent Feature Aggregation (MCRFA) module to extract one recent feature-multi complete features relation.
In addition, we organize them in the overall hierarchical-structured model to predict intention action. Our model outperforms the baselines by widely comparing it with state-of-the-art baselines on two public datasets (EPIC-Kitchens and EGTEA Gaze+).

The contributions of this paper are as follows: 1) our model innovatively employs a differentiated processing strategy for global (complete) and local (recent) features, considering their respective attributes and the temporal scale information they carry. This unique method offers a new perspective to address action anticipation.
2) We propose a GFL mechanism that facilitates interactive updates between recent and complete features through guidance, feedback, attention, residual connections, and other ideas. This GFL mechanism effectively enhances feature representation ability by enabling effective information interaction. 3) Our model adopts the concepts of global-local encoding and multitemporal scale encoding to extract temporal contexts. These approaches contribute to capturing comprehensive temporal information and improving the overall performance of the model.

\vspace{-3mm}
\section{Related work}
\vspace{-2mm}

\subsection{Intention action anticipation}
In intention action anticipation, significant advancements have been made using various approaches. Furnari \emph{et al.} \cite{furnari2019} introduced the Rolling-Unrolling Long Short-Term Memory (RU-LSTM), a novel architecture featuring two distinct LSTMs for encoding past events and predicting future actions. Building upon RU-LSTM, Osman \emph{et al.} \cite{2021SlowFast} developed the SlowFast RU-LSTM model, which processes video inputs at varying temporal scales using two branches with different frame rates.
The Ego-OMG model \cite{Dessalene2021} integrates a two-stream approach using the contact anticipation network for state sequence extraction from egocentric videos, followed by graph convolutional network layers and LSTM for further processing.

Diverging from LSTM-based methods, Sener \emph{et al.} \cite{sener2020} proposed a temporal aggregate model using coupled coupled attention mechanisms for long-range video understanding. Zatsarynna \emph{et al.} \cite{Zatsarynna2021MultiModalTC} introduced a multimodal architecture based on temporal convolutions, incorporating a novel fusion mechanism to capture interactions across RGB, FLOW, and OBJ modalities for improved multimodal integration and representation. Zheng \emph{et al.} \cite{10.1145/3544493} employ a teacher network to learn enhanced representations by considering future unobserved segments, and a student network to learn the teacher's outputs for action anticipation.

\vspace{-3mm}
\subsection{Human-object interaction inference}
\vspace{-0.5mm}
Human-object interaction (HOI) detection \cite{Cheng20231827, Yang20223853, Chiou_2021} and prediction \cite{9352754} are necessary for understanding human intention. 
The essential difference is that the intention action proposed in our work emphasizes predicting the action with the highest relevant probability for the task a person performs in a continuous process. However, HOI primarily focuses on detecting all possible actions in a specific space instead of a continuous procedure. Therefore, multiple HOIs are required to be detected in the HOI detection task, whereas only a single intent action with the highest probability needs to be predicted in our work. 
The methodologies for HOI detection and prediction are described below.

In the past, HOI detection has been popular, and researchers have proposed various valuable methods. For instance, Chao \emph{et al.} \cite{8354152} introduced a multistream model comprising human, object, and pairwise streams for detailed HOI analysis. Similarly, Fang \emph{et al.} \cite{2018Pairwise} developed a pairwise body-part attention model, concentrating on crucial body parts to enhance interaction understanding. 

Recently, the focus has shifted toward HOI prediction using neural networks, highlighting the dynamics of anticipated HOIs. Nagarajan \emph{et al.} \cite{Nagarajan2019} explored interaction hotspot mapping from videos, whereas Liu \emph{et al.} \cite{Liu2020HI} emphasized future hand motions as motor attention in their deep learning model. Furthermore, recent advancements include the Object-Centric Transformer (OCT) model proposed by researchers \cite{Liu2022JointHM}, which uses self-attention mechanisms for hand-object interaction reasoning and probabilistic frameworks for future trajectory and hotspot prediction. 

\vspace{-1mm}
\section{Method}
This section describes our proposed model for action anticipation. It is organized as follows: Section \ref{feature_extraction} depicts the feature extraction of a video sequence $\boldsymbol{V}$, Section \ref{Guide-Feedback-Loop-for-c1} and \ref{Complete-Recent-aggreation-for-c1-c2-c3} explain the details of modeling the context relations between complete and recent information of $\boldsymbol{V}$, and Section \ref{overall_architecture} introduces the hierarchical complete-recent fusion.

\vspace{-2mm}
\subsection{Feature extraction}
\label{feature_extraction}
Given the input video sequence $\boldsymbol{V}$, we extract two sets of features, including the complete feature set $\boldsymbol{C}$ and the recent feature set $\boldsymbol{R}$.
$\boldsymbol{C}$ comprises three feature scales:
\begin{equation}
	\boldsymbol{C}=\left\{\boldsymbol{C}_1, \boldsymbol{C}_2, \boldsymbol{C}_3\right\},
	\label{eq:complete_feature}
\vspace{-1mm}
\end{equation}
where $\boldsymbol{C}_1$, $\boldsymbol{C}_2$, and $\boldsymbol{C}_3$ correspond to the complete features when $\boldsymbol{V}$ is divided into different fragment scales. The motivation is to strengthen the representation of the complete video sequence. Furthermore, this method can handle videos with different durations.

Assuming that $\boldsymbol{C}_j$ ($j=1,2,3$) refers to the complete feature when $\boldsymbol{V}$ is divided into $n(j)$ fragments, then $\boldsymbol{C}_j$ is an $n(j)$ dimension feature:
\begin{equation}
\setlength{\abovedisplayskip}{3pt}
\setlength{\belowdisplayskip}{3pt}
	{\boldsymbol{C}_j}=\left\{c_m|m=1,2,\ldots, n(j) \right\},
	\label{cc}
\end{equation}
where each element $c_m$ is obtained by applying the max-pooling operation to the features of the frames belonging to $m_{th}$ fragments. $n(j)$ varies with the value of $j$. 

$\boldsymbol{R}$ comprises four feature scales:
\begin{equation}
\setlength{\abovedisplayskip}{3pt}
\setlength{\belowdisplayskip}{3pt}
	\boldsymbol{R}=\left\{\boldsymbol{R}_1, \boldsymbol{R}_2, \boldsymbol{R}_3, \boldsymbol{R}_4\right\},
	\label{eq:four_recent_feature}
\end{equation}
where $\boldsymbol{R}_1$, $\boldsymbol{R}_2$, $\boldsymbol{R}_3$ and $\boldsymbol{R}_4$ denote the recent features when the $\Delta t$  parameters are set as different durations.
$\boldsymbol{R}_i$ ($i=1,2,3,$ or $4$) is defined as follows:
\begin{equation}
\setlength{\abovedisplayskip}{3pt}
\setlength{\belowdisplayskip}{3pt}
	\boldsymbol{R}_i=\left\{r_m|m=1,2\right\},
	\label{eq:recent_features}
\end{equation}
where each element $r_{m}$ is obtained by applying the max-pooling operation on the features of the frames belonging to two fragments. The feature extraction method is the same as that for $c_m$ in Equation~\ref{cc}. Max-pooling aims to aggregate and compress information from fragments efficiently. This operation allows our model to capture the most prominent features within each fragment, thereby retaining critical information and enhancing model efficiency.

For fairness, we follow the method adopted in previous studies \cite{furnari2019,sener2020, Zatsarynna2021MultiModalTC, 2021TransAction} to extract the features for every video frame. Each frame has three modalities: RGB (i.e., image feature), FLOW (i.e., optical flow feature), and OBJ (i.e., object feature).

\begin{figure}[t]
	\centering
	\includegraphics[width=0.6\textwidth]{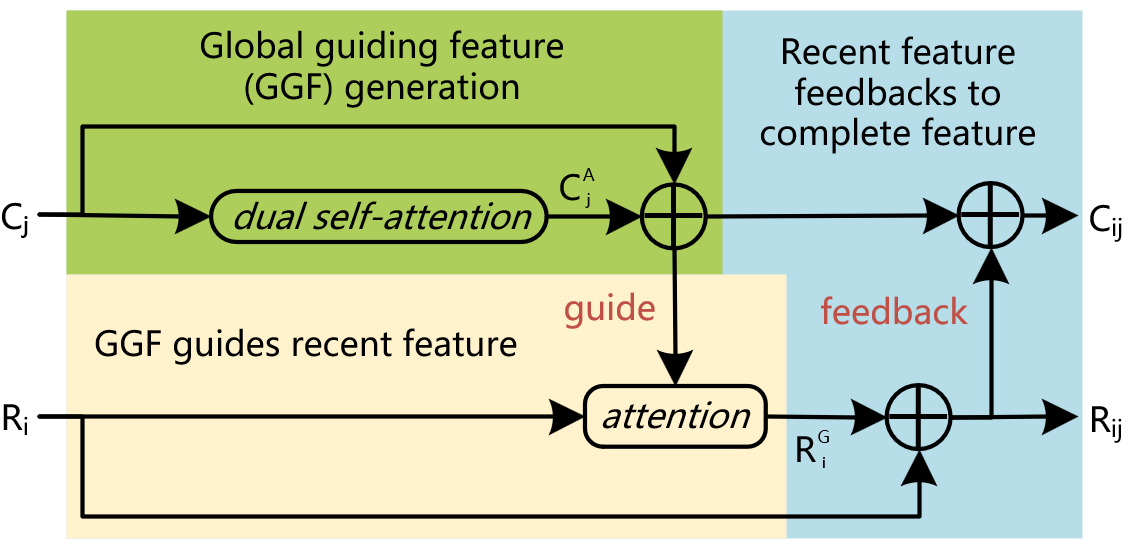}
  \vspace{-3mm}
	\caption{Overview of Guide-Feedback Loop (GFL) mechanism. GFL comprises three stages: 1) complete feature is updated to generate a global guiding feature ($\boldsymbol{GGF}$); 2) global guiding feature guides the recent feature; 3) guided recent feature feeds back to the updated complete feature.}
	\label{fig_GFL}
 \vspace{-3mm}
\end{figure}
\vspace{-4mm}
\subsection{Guide-Feedback Loop mechanism}
\label{Guide-Feedback-Loop-for-c1}
We model the relation between one complete feature $\boldsymbol{C}_j$ and one recent feature $\boldsymbol{R}_i$ by proposing the three-stage GFL (Fig.~\ref{fig_GFL}), formulated as follows:
\begin{equation}
\setlength{\abovedisplayskip}{3pt}
\setlength{\belowdisplayskip}{3pt}
	\boldsymbol{R}_{ij}, \boldsymbol{C}_{ij}= GFL(\boldsymbol{R}_i, \boldsymbol{C}_j),
	\label{AB1}
\end{equation}
where $\boldsymbol{R}_{ij}$ and $\boldsymbol{C}_{ij}$ are the results of updating $\boldsymbol{R}_i$ and $\boldsymbol{C}_j$. The three-stage of GFL is detailed as follows:

\noindent{\textit{\textbf{Stage 1 - Global guiding feature generation.}}}
This stage generates a global guiding feature ($\boldsymbol{GGF}$) to guide $\boldsymbol{R}_i$ in the second stage.
We generate $\boldsymbol{GGF}$ by designing a ``dual self-attention + residual'' structure. This structure is designed with two primary motivations.

First, the complete feature $\boldsymbol{C}_j$ carries global information. However, this information is not weighted according to its importance. Therefore, we introduce a dual self-attention mechanism to address this issue. We can assign weights to different parts of $\boldsymbol{C}_j$ by applying self-attention twice based on their significance. This weighting strategy allows us to emphasize crucial information. Consequently, the generated $\boldsymbol{GGF}$ captures the most relevant and informative aspects of $\boldsymbol{C}_j$.
Second, we incorporate a residual structure to preserve the original information in $\boldsymbol{C}_j$. The residual connection enables the flow of information from the input $\boldsymbol{C}_j$ to the final output $\boldsymbol{GGF}$. Therefore, the model can retain valuable contextual cues $\boldsymbol{C}_j$ provides while taking advantage of the refined and weighted representation obtained through the dual self-attention mechanism.

The dual self-attention mechanism is applied on $\boldsymbol{C}_j$ to obtain the attention complete feature $\boldsymbol{C}_j^A$. The formula is expressed as follows:
\begin{equation}
\setlength{\abovedisplayskip}{3pt}
\setlength{\belowdisplayskip}{3pt}
	\boldsymbol{C}_j^A= Soft(Conv(\boldsymbol{C}_j))*\boldsymbol{C}_j*Soft(Conv(\boldsymbol{C}_j)),
	\label{k1}
\end{equation}
where $Conv$ is the convolution, $Soft$ represents the softmax function, and $*$ denotes matrix multiplication. Then, $\boldsymbol{GGF}$ is computed as follows:
\begin{equation}
\setlength{\abovedisplayskip}{3pt}
\setlength{\belowdisplayskip}{3pt}
	\boldsymbol{GGF} = \boldsymbol{C}_j+Drop(Conv(ReLU(LN(Conv(\boldsymbol{C}_j^A))))),
	\label{a1}
\end{equation}
where $LN$ denotes LayerNorm and $Drop$ is the dropout. 
The ``dual self-attention + residual'' mechanism effectively addresses the twofold challenge of weighting the importance of information in $\boldsymbol{C}_j$ and maintaining the original information during $\boldsymbol{GGF}$ generation. This allows us to highlight crucial features and retain the essential context, ultimately enhancing the guidance provided to the recent feature $\boldsymbol{R}_i$ in the subsequent stage of the model.

\noindent{\textit{\textbf{Stage 2 - GGF guides recent feature.}}}
This stage aims to update the recent feature $\boldsymbol{R}_i$ under the guidance of $\boldsymbol{GGF}$. To this end, we use the attention mechanism to incorporate the information of the complete feature into the recent feature. The attention mechanism enables us to selectively focus on the relevant parts of the complete feature and integrate their information into the recent feature. 
By weighting the importance of regions or elements of features,
we can enhance the discriminative power and contextual understanding of the recent feature. Mathematically, the transformation can be formulated as follows:
\begin{equation}
\setlength{\abovedisplayskip}{3pt}
\setlength{\belowdisplayskip}{3pt}
	\boldsymbol{R}_i^G= Soft(Conv(\boldsymbol{R}_i))*\boldsymbol{GGF}*Soft(Conv(\boldsymbol{GGF})),
	\label{k2}
\end{equation}
where the guided recent feature $\boldsymbol{R}_i^G$ captures the original characteristics of $\boldsymbol{R}_i$ and the enriched information derived from the complete feature. The purpose of applying the softmax operation is to transform the resulting values into a probability distribution that represents the importance weights of each location. Multiplying $\boldsymbol{GGF}$ with $Soft(Conv\allowbreak(\boldsymbol{GGF}))$ to highlights crucial information by weighting important areas or elements in $\boldsymbol{GGF}$. The purpose of multiplying $\boldsymbol{GGF}*Soft(Conv(\boldsymbol{GGF}))$ with $Soft(Conv(\boldsymbol{R}_i))$ is to adjust the importance weights of different positions in the recent feature. The information is effectively combined by multiplying the convolution of the weighted recent feature with the complete feature.

\noindent{\textit{\textbf{Stage 3 - Recent feature feedbacks to complete feature.}}}
At this stage, we first update $\boldsymbol{R}_i^G$ to $\boldsymbol{R}_{ij}$ at first. The formula is as follows:
\begin{equation}
\setlength{\abovedisplayskip}{3pt}
\setlength{\belowdisplayskip}{3pt}
	\boldsymbol{R}_{ij}= Drop(Conv(ReLU(LN(Conv(\boldsymbol{R}_i^G)))))+\boldsymbol{R}_i,
	\label{o2}
\end{equation}
where $\boldsymbol{R}_{ij}$ is the updated recent feature. Then, $\boldsymbol{R}_{ij}$ integrates the information into the complete feature through feedback. This integration allows for the exchange of information between the global long-term context represented by $\boldsymbol{C}_j$ and the recently enhanced features reflected in $\boldsymbol{R}_{ij}$. The updating of $\boldsymbol{C}_j$ as $\boldsymbol{C}_{ij}$ is computed as follows:
\begin{equation}
\setlength{\abovedisplayskip}{3pt}
\setlength{\belowdisplayskip}{3pt}
	\boldsymbol{C}_{ij}= \boldsymbol{R}_{ij}+\boldsymbol{GGF},
	\label{o1}
\end{equation}
where $\boldsymbol{C}_{ij}$ contains the global long-term information and the recent information. The above operations realize the interactive updating of the two features.

The GFL mechanism is interactive and fundamentally different from simple addition and connection mechanisms. Through the ideas of guidance, feedback, attention, and residual, it can better realize the interactive fusion of one complete feature and one recent feature. 

\begin{figure}[t]
	\centering
	\includegraphics[width=0.8\textwidth]{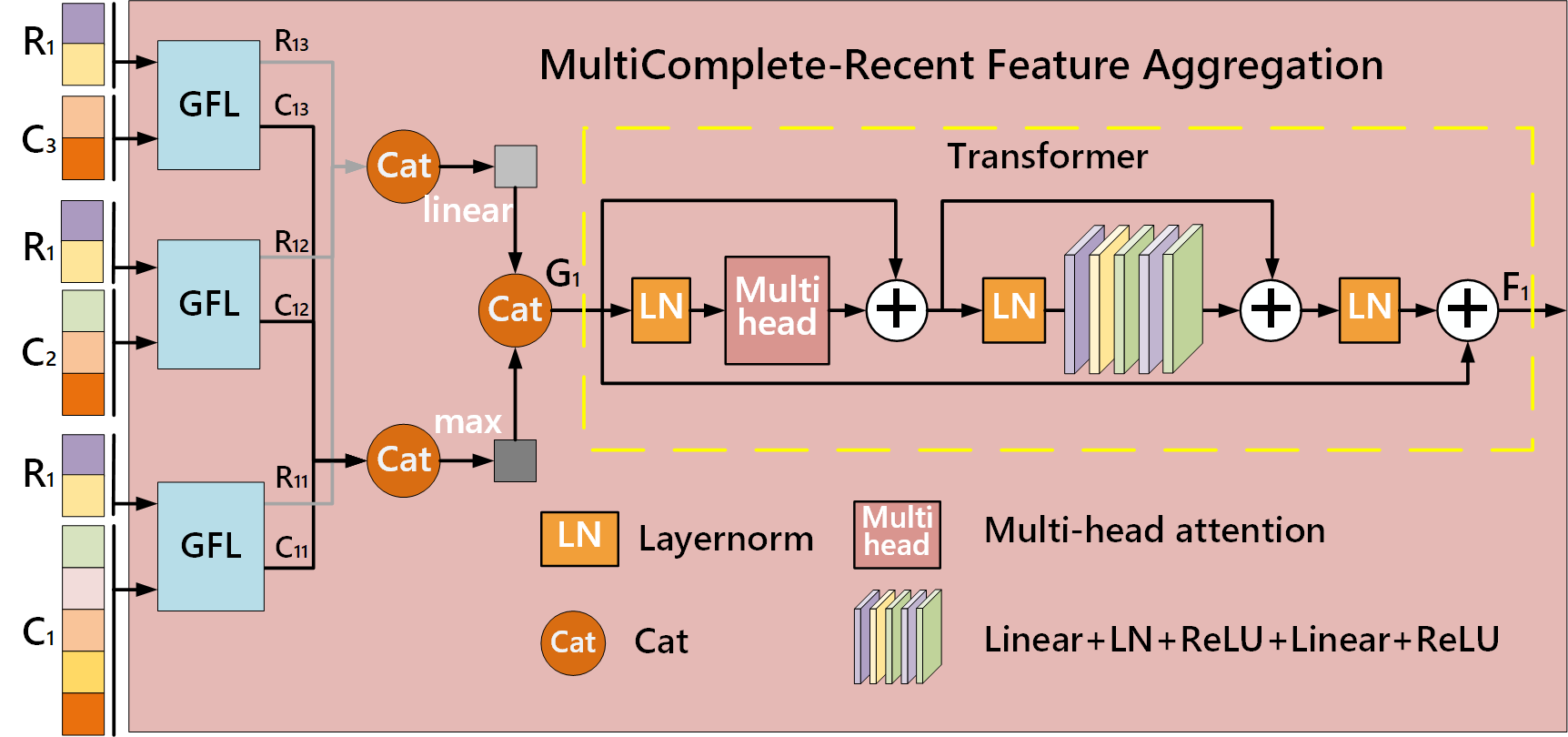}
	\caption{Overview of MultiComplete-Recent Feature Aggregation (MCRFA) module. The MCRFA module models one recent feature with multiple complete features.}
	\label{fig_Net}
	\vspace{-0.4cm}
\end{figure}

\vspace{-3mm}
\subsection{MultiComplete-Recent Feature Aggregation}
\label{Complete-Recent-aggreation-for-c1-c2-c3}
We propose the Transformer-based MultiComplete-Recent Feature Aggregation (MCRFA) to model the relation of one recent feature $\boldsymbol{R}_i$ with all complete features (i.e., $\boldsymbol{C}_1$, $\boldsymbol{C}_2$ and $\boldsymbol{C}_3$). The formula is as follows:
\begin{equation}
\setlength{\abovedisplayskip}{3pt}
\setlength{\belowdisplayskip}{3pt}
	\boldsymbol{F}_i= MCRFA(\boldsymbol{C}_1,\boldsymbol{R}_i,\boldsymbol{C}_2,\boldsymbol{R}_i,\boldsymbol{C}_3,\boldsymbol{R}_i), i=1,2,3,4,
\end{equation}
where $\boldsymbol{F}_i$ is the output of the MCRFA. To better illustrate this process, we use $\boldsymbol{R}_1$ as an example, as shown in Fig.~\ref{fig_Net}.
First, MCRFA uses three GFL modules to obtain the updated recent and complete features, which can take full advantage of GFL to capture information from different periods, improving feature expressibility and predictive performance.
\begin{equation}
\setlength{\abovedisplayskip}{3pt}
\setlength{\belowdisplayskip}{3pt}
	\left\{\begin{array}{l}
		\boldsymbol{R}_{11},\boldsymbol{C}_{11}= GFL(\boldsymbol{R}_1,\boldsymbol{C}_1)\\
		\boldsymbol{R}_{12},\boldsymbol{C}_{12}= GFL(\boldsymbol{R}_1,\boldsymbol{C}_2)\\
		\boldsymbol{R}_{13},\boldsymbol{C}_{13}= GFL(\boldsymbol{R}_1,\boldsymbol{C}_3)\\
	\end{array}\right.,
\end{equation}
where $\boldsymbol{R}_{1j}$ represents the updating of $\boldsymbol{R}_1$ under the guidance of $\boldsymbol{C}_j$, and $\boldsymbol{C}_{1j}$ represents the updating of $\boldsymbol{C}_j$ containing the information of $\boldsymbol{R}_1$.

Then, we aggregate $\{\boldsymbol{R}_{1j}|j=1,2,3\}$ and $\{\boldsymbol{C}_{1j}|j=1,2,3\}$ to obtain the aggregated information $\boldsymbol{G}_1$. The formula is as follows:
\begin{equation}
\setlength{\abovedisplayskip}{3pt}
\setlength{\belowdisplayskip}{3pt}
		\boldsymbol{G}_1 = {\it Cat}(M(Cat(\boldsymbol{C}_{11},\boldsymbol{C}_{12},\boldsymbol{C}_{13}),L(Cat(\boldsymbol{R}_{11},\boldsymbol{R}_{12},\boldsymbol{R}_{13})))),
		\label{o}
\end{equation}
where $Cat$ represents a concatenation operation, $M$ denotes the operation of taking the maximum value, and $L$ is the linear transformation, unlike many existing methods that only use the attention mechanism for aggregation. 

The aggregated feature $\boldsymbol{G}_1$ is further fed into the Transformer to strengthen the aggregated feature $\boldsymbol{F}_1$. We designed this because the Transformer techniques allow the model to effectively capture global context and complex feature relationships and enhance feature representations.
\begin{equation}
\setlength{\abovedisplayskip}{3pt}
\setlength{\belowdisplayskip}{3pt}
	\boldsymbol{F}_1 = Transformer(\boldsymbol{G}_1),
	\label{f}
\end{equation}
where $\boldsymbol{F}_1$ models the relationship between $\boldsymbol{R}_1$ and $\{\boldsymbol{C}_j|j=1,2,3\}$.

\begin{figure}[!t]
	\centering
	\includegraphics[width=0.7\textwidth]{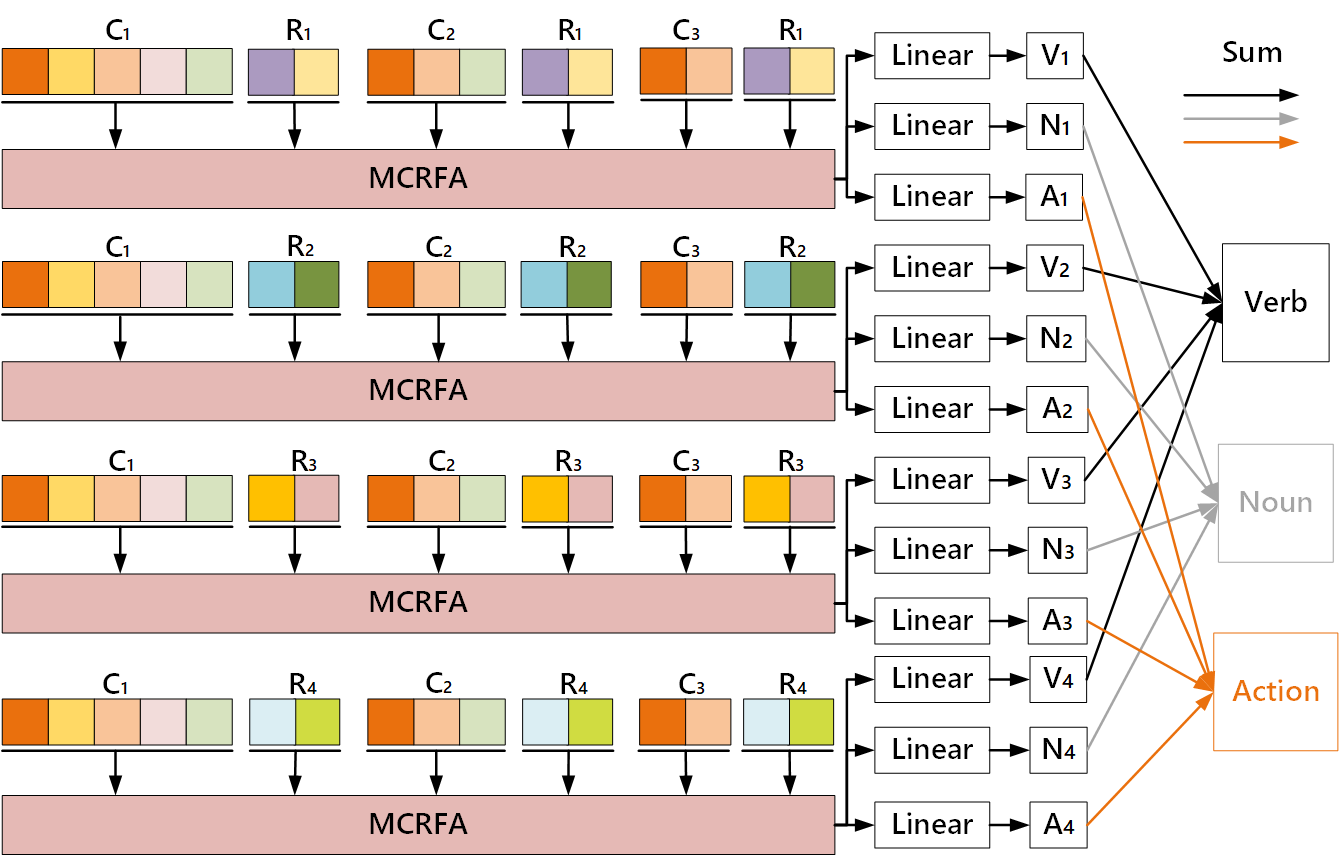}
 	\vspace{-4mm}
	\caption{Hierarchical complete-recent fusion prediction. `V' represents a verb, `N' denotes a verb, and `A' expresses an action.}
	\label{fig_all}
	\vspace{-6mm}
\end{figure}
\vspace{-4mm}
\subsection{Hierarchical Complete-Recent Fusion}
\label{overall_architecture}
Section \ref{Complete-Recent-aggreation-for-c1-c2-c3} have modeled one recent feature (i.e., $\boldsymbol{R}_1$) with all complete features (i.e., $\boldsymbol{C}_1$, $\boldsymbol{C}_2$, and $\boldsymbol{C}_3$). This section explores the relations of all recent and complete features to construct our overall model architecture, which is illustrated in Fig.~\ref{fig_all}. 

By adopting four MCRFA modules, $\{\boldsymbol{F}_i|i=1,2,3,4\}$ is obtained.
Then, linear projections are applied on $\boldsymbol{F}_i$ to acquire four groups of intention verb, noun, and action $\left\{\boldsymbol{V}_i, \boldsymbol{N}_i, \boldsymbol{A}_i \right\}_{i=1}^4$.
\begin{equation}
\setlength{\abovedisplayskip}{3pt}
\setlength{\belowdisplayskip}{3pt}
	\boldsymbol{V}_i, \boldsymbol{N}_i, \boldsymbol{A}_i = Linears(\boldsymbol{F}_i),\\
	\label{eq:intention_triplet}
\end{equation}
where $Linears$ refer to the multiple linear transformation layers mapping $\boldsymbol{F}_i$ to $\boldsymbol{V}_i$, $\boldsymbol{N}_i$, and $\boldsymbol{A}_i$. The final intention results are obtained by summarizing the predicted verbs, nouns, and actions in Equation~\ref{eq:intention_triplet}.  
\begin{equation}
\setlength{\abovedisplayskip}{3pt}
\setlength{\belowdisplayskip}{3pt}
	\left\{\begin{array}{l}
		\boldsymbol{Verb} = \boldsymbol{V}_1 + \boldsymbol{V}_2 + \boldsymbol{V}_3 + \boldsymbol{V}_4\\
		\boldsymbol{Noun} = \boldsymbol{N}_1 + \boldsymbol{N}_2 + \boldsymbol{N}_3 + \boldsymbol{N}_4\\
		\boldsymbol{Action} = \boldsymbol{A}_1 + \boldsymbol{A}_2 + \boldsymbol{A}_3 + \boldsymbol{A}_4\\
	\end{array}\right.,
\end{equation}
where $\boldsymbol{Verb}$, $\boldsymbol{Noun}$, and $\boldsymbol{Action}$ are the final intention results.
So far, our descriptions are based on one-modal data. To predict intention using multi-modal data, a typical late-fusion manner is used. The late-fusion mechanism averagely fuses the prediction results of different modalities to produce a final prediction. In particular, after each modality has made its independent prediction, these predictions are combined through a weighted average to arrive at the final decision.

\vspace{-3mm}
\section{Experiments}
\vspace{-1mm}
\subsection{Experiment setup}
\noindent{\textit{\textbf{Datasets.}}}
We evaluate our method on two large-scale egocentric datasets: EPIC-Kitchens (EPIC) \cite{Damen2018ScalingEV} and EGTEA Gaze+ (EGTEA) \cite{Li2018InTE}.
EPIC comprises 55 hours of recordings capturing all daily activities in kitchens and the dataset has 2,513 unique action classes, 125 verb classes, and 352 noun classes. 
For EPIC, some other works (e.g. \cite{sener2020, 2021SlowFast}) and we use the dataset split proposed in \cite{furnari2019} for a fair comparison. 
EGTEA consists of 10,321 annotated video clips with 19 verb classes, 51 noun classes, and 106 action classes. The authors \cite{Li2018InTE} of EGTEA have split the dataset into three parts. We report the average performance of the three parts. 

\noindent{\textit{\textbf{Metrics.}}}
Based on the study suggested by \cite{furnari2019,sener2020}, we adopt top-5 accuracy and mean top-5 recall to evaluate the results of our experiments. Top-5 Accuracy measures whether the correct category label is included among the top five results predicted by the model. A prediction is considered correct if the correct label is within the model’s top five predicted categories. The Mean Top-5 Recall (M.Top-5 Rec) is a class-aware metric. For a given category $c$, the Top-5 Recall is defined as the proportion of samples whose true category is $c$ and that includes category $c$ among the top five most likely predicted actions. The Mean Top-5 Recall is the average of the Top-5 Recall across all categories.
According to \cite{0Leveraging}, the top-1 accuracy measure is not always appropriate for evaluating multilabel problems; therefore, in this paper, $k=5$.

\noindent{\textit{\textbf{Implementation Details.}}}
Following the method adopted in previous studies \cite{furnari2019,sener2020,2021TransAction}, 
we use the image features extracted by \cite{furnari2019}.
The RGB and flow features are extracted using TSN \cite{Wang2016TemporalSN}. A Faster Region-based Convolutional Neural Network (Faster R-CNN) \cite{Ren2015FasterRT} object detector with the ResNet-101 backbone extracts the object features. 

\vspace{-0.5mm}
\textbf{1)} During the training, the optimizer is Adam. The learning rate is set to 1e-4, decay is performed every eight epochs, and the learning rate is multiplied by 0.1 for each decay. The batch size is 10. The cross-entropy loss is used to compute the losses of the predicted four intention action groups $\left\{\boldsymbol{V}_i, \boldsymbol{N}_i, \boldsymbol{A}_i \right\}$ in Equation~\ref{eq:intention_triplet} and their corresponding labels, and summarize them as the total loss.
\textbf{2)} In Equation~\ref{cc}, $n(1)=2$, $n(2)=3$, and $n(3)=5$ on EPIC-Kitchens and $n(1)=2$, $n(2)=4$, and $n(3)=8$ on EGTEA Gaze+.
In Equation~\ref{eq:four_recent_feature}, the parameters of $\Delta t$ corresponding to $\left\{\boldsymbol{R}_1, \boldsymbol{R}_2, \boldsymbol{R}_3, \boldsymbol{R}_4 \right\}$ are $\left\{1.6,1.2,0.8,0.4\right\}$.
\textbf{3)} In Section 3.3, the number of heads and layers in the Transformer on EPIC are 5 and 1, respectively. The number of heads and layers in the Transformer on EGTEA are 2 and 1, respectively. 

\begin{table*}[t]
	\caption{The comparison results with baselines (anticipation time $\tau$ is 1s).}
	\centering
	\resizebox{12cm}{!}{
		\begin{tabular}{rcccccccccccccc}
			\bottomrule[1pt]
			\multirow{3}*{Method}&\multicolumn{7}{c}{EPIC-Kitchens}&\multicolumn{7}{c}{EGTEA Gaze+}\\
			\cline{2-15}
			&\multicolumn{3}{c}{Top-5 Acc}&\multicolumn{3}{c}{ M.Top-5 Rec}&\multirow{2}*{Avg.}&\multicolumn{3}{c}{Top-5 Acc} & \multicolumn{3}{c}{ M.Top-5 Rec}&\multirow{2}*{Avg.}\\
			\cline{2-7}
			\cline{9-14}
			&VERB&NOUN&ACT&VERB&NOUN&ACT&&VERB&NOUN&ACT&VERB&NOUN&ACT&\\
			\cline{1-15}
			ED \cite{2017RED}&75.46&42.96&25.75&41.77&42.59&10.97&39.92&86.79&64.35&50.22&69.66&56.62&42.74&61.73\\
			
			FN \cite{2018Modeling}&74.84&40.87&26.27&35.30&37.77&6.64&36.95&91.05&71.64&60.12&76.73&63.59&49.82&68.83\\
	
		RULSTM \cite{furnari2019} &79.55&51.79&\underline{35.32}&43.72&49.90&15.10&45.90&\textbf{93.11}&77.48&66.40&82.07&73.30&58.64&75.17\\
			TAB \cite{sener2020} &79.47&\underline{51.93}&34.60&\underline{44.15}&\underline{51.88}&\underline{16.17}&\underline{46.37}&92.84&\underline{78.58}&67.51&\underline{83.12}&\underline{75.17}&\underline{62.46}&\underline{76.61}\\	HA \cite{2021TransAction}&73.94&41.29&25.18&35.87&36.39&14.64&37.89&86.06&71.46&59.89&78.90&69.89&58.22&70.74\\
			IRNN \cite{2021LearningAnticipate}&\underline{79.70}&50.20&33.20&-&-&-&-&-&-&66.71&-&-&-&-\\

  SRL \cite{Qi2021SelfRegulatedLF}  &78.90&47.65&31.68& 42.83& 47.64& 13.24&41.45&\underline{93.03}&78.09&\underline{68.01}&79.67&72.67&60.86&75.39\\
  OCT \cite{Liu2022JointHM}&73.90&45.90&24.40&-&-&-&-&-&-&-&-&-&-&-\\
   S+D \cite{Furnari2023}& 74.33 &45.42&29.23&-&-&-&-& 90.83&76.71&66.00&-&-&-&-\\
			\midrule[0.5pt]
			\textbf{Ours}&\textbf{80.26}&\textbf{53.10}&\textbf{35.45}&\textbf{44.84}&\textbf{53.43}&\textbf{17.14}&\textbf{47.37}&\textbf{93.11}&\textbf{79.20}&\textbf{69.03}&\textbf{83.16}&\textbf{75.22}&\textbf{64.65}&\textbf{77.40}\\
			\bottomrule[1pt]
		\end{tabular}
	}
	\label{epic}
 \vspace{-4mm}
\end{table*}

\subsection{Comparison experiments}
\noindent{\textit{\textbf{Quantitative results.}}}
We evaluate the performance of our model and baselines for verb, noun, and action predictions on EPIC and EGTEA, and the results are shown in Table~\ref{epic}. In the table, $Avg.$ is the average value of the metric results for each method. 
Intention action anticipation in egocentric videos is extremely challenging because the number of possible intention actions is enormous. In addition, human activities are complex, dynamic, and random, sometimes releasing misleading signals for intention analysis. For example, when a human’s hand is near an object, it is unclear whether the human will approach or pull away from the object; therefore, it is challenging for the model to accurately select the correct verb. Consequently, the overall performance of the models proposed recently typically increased slightly. 
For example, the average performance of TAB \cite{sener2020} is 1.02\% higher than that of RU-LSTM \cite{furnari2019} on EPIC and 1.92\% on EGTEA. 
Compared with the second-best model (i.e., TAB \cite{sener2020}), our model obtains an average performance improvement of 2.16\% on EPIC and 1.03\% on EGTEA.

\subsection{Diagnostic study}
\noindent{\textit{\textbf{The effect of GFL.}}}
Our investigations into the GFL module, a core component of our model, as discussed in Section \ref{Guide-Feedback-Loop-for-c1}, reveal its significant impact on performance. We assessed its importance by conducting experiments in which GFL was removed, and the results are detailed in Table~\ref{GFLepic}. These results demonstrate the model's superior performance with GFL, underlining the module's effectiveness in modeling the interplay between current and complete information. The experiments highlight the GFL module's vital role in capturing temporal dependencies and enhancing our model's representational capabilities, with its removal resulting in a marked performance decline.
\begin{table*}[t]
	\caption{Ablation results of the GFL module.}
	\centering
	\resizebox{12cm}{!}{
		\begin{tabular}{ccccccccccccc}
			\toprule[1pt]
			\multirow{3}*{Module}&\multicolumn{6}{c}{EPIC-Kitchens}&\multicolumn{6}{c}{EGTEA Gaze+}\\
			\cline{2-13}
			&\multicolumn{3}{c}{Top-5 Acc}& \multicolumn{3}{c}{M.Top-5 Rec}&\multicolumn{3}{c}{Top-5 Acc}& \multicolumn{3}{c}{M.Top-5 Rec}\\
			\cline{2-13}
			&VERB&NOUN&ACT&VERB&NOUN&ACT&VERB&NOUN&ACT&VERB&NOUN&ACT\\
			\cline{1-13}
			w/o \textbf{GFL}&71.10&34.03&23.90&31.19&29.93&5.47&89.94&73.63&62.54&76.97&63.34&53.64\\
			\midrule[0.5pt]
			with \textbf{GFL}&\textbf{80.26}&\textbf{53.10}&\textbf{35.45}&\textbf{44.84}&\textbf{53.43}&\textbf{17.14}&\textbf{93.11}&\textbf{79.20}&\textbf{69.03}&\textbf{83.16}&\textbf{75.22}&\textbf{64.65}\\
			\bottomrule[1pt]
		\end{tabular}
	}
	\label{GFLepic}
 \vspace{-3mm}
\end{table*}

\begin{table*}[t]
	\caption{Ablation results of guide and feedback.}
	\centering
	\resizebox{12cm}{!}{
		\begin{tabular}{ccccccccccccc}
			\toprule[1pt]
			\multirow{3}*{Module}&\multicolumn{6}{c}{EPIC-Kitchens}&\multicolumn{6}{c}{EGTEA Gaze+}\\
			\cline{2-13}
			&\multicolumn{3}{c}{Top-5 Acc}& \multicolumn{3}{c}{M.Top-5 Rec}&\multicolumn{3}{c}{Top-5 Acc}& \multicolumn{3}{c}{M.Top-5 Rec}\\
			\cline{2-13}
			&VERB&NOUN&ACT&VERB&NOUN&ACT&VERB&NOUN&ACT&VERB&NOUN&ACT\\
			\cline{1-13}
			None&79.70&52.12&34.74&44.04&53.08&15.78&92.39&78.56&68.26&81.69&74.33&63.47\\
			G&79.98&52.32&35.07&44.03&52.67&15.99&92.74&78.70&68.38&81.87&74.58&64.06\\
			F&79.92&52.92&35.19&44.33&53.15&15.83&92.75&78.80&68.01&81.73&74.68&63.82\\
			\midrule[0.5pt]
			\textbf{G+F}&\textbf{80.26}&\textbf{53.10}&\textbf{35.45}&\textbf{44.84}&\textbf{53.43}&\textbf{17.14}&\textbf{93.11}&\textbf{79.20}&\textbf{69.03}&\textbf{83.16}&\textbf{75.22}&\textbf{64.65}\\
			\bottomrule[1pt]
		\end{tabular}
	}
	\label{guide_feed}
 \vspace{-3mm}
\end{table*}

\noindent{\textit{\textbf{The effect of guide and feedback.}}} 
Our analysis of the GFL module involved experiments with three configurations: no loop ($None$), only guidance ($G$), only feedback ($F$), and complete GFL mechanism ($G+F$). Results. As detailed in Table~\ref{guide_feed}, the results indicate that the model performs optimally when the guidance and feedback components are active. While $G$ and $F$ individually outperform the absence of a loop, they are less effective than the integrated $G+F$ mechanism, confirming the efficacy of the GFL mechanism in our model.

The success of this mechanism can be attributed to two fundamental processes: the guidance process, which enriches recent features with contextual information from complete features while retaining original information, and the feedback process, where recent features enhance the complete features with up-to-date data. Collectively, these processes enable a more comprehensive understanding of human intention, enhancing the overall model performance.

\begin{figure}[!t]
	\centering
	\includegraphics[width=0.8\textwidth]{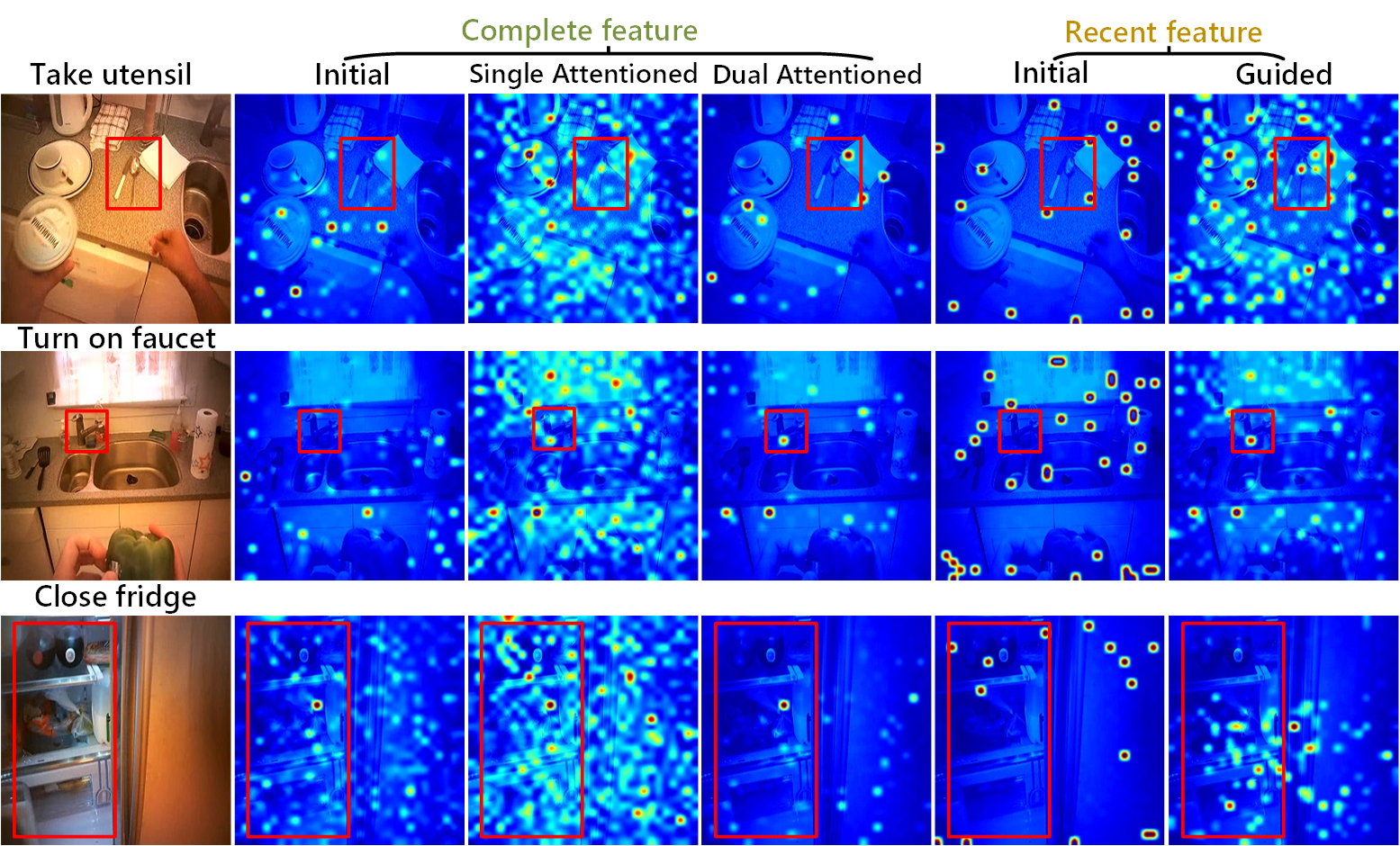}
  \vspace{-4mm}
	\caption{Attention visualization of the complete and recent features. Red boxes indicate ground truth regions. Initial represents the initial features. 
Single Attentioned denotes the results for which only a single self-attention mechanism is used (i.e., Equation~\ref{k1} converts to $\boldsymbol{K_1}= Soft(Conv(\boldsymbol{C}_1))*\boldsymbol{C}_1)$).
Dual Attentioned denotes the results generated by the dual self-attention mechanism (Equation~\ref{k1}). 
Guided denotes the results of recent features guided by $\boldsymbol{GGF}$ (Equation~\ref{k2})}
	\label{attention}
  \vspace{-6mm}
\end{figure}

\noindent{\textit{\textbf{The visualization of features in GFL.}}} We further verify the effectiveness of the GFL by visualizing the attention of complete features and recent features in different stages of GFL. The results are shown in Fig.~\ref{attention}.

The initial complete features show almost no high-probability attention spots inside the red boxes.
The initial complete features pay more attention to the area of attention at the previous moment, such as the initial scenarios in the first row of the image. Before approaching the utensil, people will first approach the kitchen counter where it is placed and pay attention to it. Most other initial scenarios are similar.
After training, the complete and recent features moved their high-probability attention spots toward the red boxes. 
Specifically, through a single self-attention mechanism, although more attention hotspots of complete features are inside the boxes, many are outside the boxes, and the attention points are scattered. After the dual self-attention mechanism, the attention hotspots of the dual attention complete features are concentrated in the red boxes and are no longer spread over a large area (column 5 in Fig.~\ref{attention}).
For recent features, the attention is disorderly distributed in different locations of initial features through the guiding procedure (Equation~\ref{k2}), and the attention tends to converge inside the boxes.

\begin{table*}[!h]
	\caption{Ablation results of complete features and recent features.}
	\centering
	\resizebox{12cm}{!}{
		\begin{tabular}{ccccccccccccccc}
			\toprule[1pt]
			\multirow{3}*{Module}&\multicolumn{7}{c}{EPIC-Kitchens}&\multicolumn{7}{c}{EGTEA Gaze+}\\
			\cline{2-15}
			&\multicolumn{3}{c}{Top-5 Acc}& \multicolumn{3}{c}{M.Top-5 Rec}&\multirow{2}*{Avg.}&\multicolumn{3}{c}{Top-5 Acc}& \multicolumn{3}{c}{M.Top-5 Rec}&\multirow{2}*{Avg.}\\
			\cline{2-7}
			\cline{9-14}
   &VERB&NOUN&ACT&VERB&NOUN&ACT&&VERB&NOUN&ACT&VERB&NOUN&ACT&\\
			\cline{1-15}
			C&80.06&52.44&34.34&44.06&51.79&15.33&46.34&92.25&77.28&65.99&78.62&73.12&61.95&74.87\\
			R&79.68&50.84&33.73&44.23&50.86&15.82&45.86&91.71&75.83&64.62&79.28&71.55&60.13&73.85\\
			\midrule[0.5pt]
			\textbf{C+R}&\textbf{80.26}&\textbf{53.10}&\textbf{35.45}&\textbf{44.84}&\textbf{53.43}&\textbf{17.14}&\textbf{47.37}&\textbf{93.11}&\textbf{79.20}&\textbf{69.03}&\textbf{83.16}&\textbf{75.22}&\textbf{64.65}&\textbf{77.40}\\
			\bottomrule[1pt]
		\end{tabular}
	}

	\label{inputepic}
	\vspace{-4mm}
\end{table*}

\noindent{\textit{\textbf{The effects of complete and recent features.}}}
The complete features contain the global information of the entire video, providing a comprehensive view and guidance for the model. Conversely, recent features focus on local information at the end of the video, capturing critical cues shortly before human actions are performed. These two features complement each other. We demonstrated the importance of these two features and their complementary nature by conducting experiments with three configurations: using only the complete feature ($C$), using only the recent feature ($R$), and combining both features ($C+R$). The results are summarized in Table ~\ref{inputepic}.
The analysis presented in Table~\ref{inputepic} reveals that the $C+R$ configuration outperforms $C$ and $R$ individually, exhibiting an average performance improvement of 1.78\% and 2.53\%, respectively. These results further emphasize that combining complete and recent features enhances the model's performance reinforcing the importance of integrating diverse features and highlighting the benefits of considering the broader video context and the specific temporal cues before human actions.

\begin{figure}[!t]
	\centering
\includegraphics[width=0.45\textwidth]{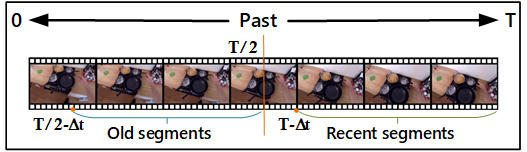}
   \vspace{-5mm}
	\caption{Definition of old and recent segments. The interval for old segments is [T/2-$\Delta t$, T/2], and the interval for recent segments is [T-$\Delta t$, T].}
	\label{fig_old_recent}
  \vspace{-4mm}
\end{figure}

\noindent{\textit{\textbf{The effect of old and recent features.}}}
As shown in Fig.~\ref{fig_old_recent}, old features are extracted from old segments, and recent features are extracted from recent segments.
Table~\ref{time} compares the recent and old features in four $\Delta t$ cases (recent features have four scales; therefore, $\Delta t$ contains four values, such as $\left\{0.4, 0.3, 0.2, 0.1 \right\}$). Overall, the performance of the model using the recent feature gains an average improvement of 4.44\% over that of the model using the old feature. Specifically, when the values of $\Delta t$ are $\left\{0.4, 0.3, 0.2, 0.1 \right\}$, $\left\{0.8, 0.6, 0.4, 0.2 \right\}$, $\left\{1.2, 0.9, 0.6, 0.3 \right\}$, and $\left\{1.6, 1.2, 0.8, 0.4 \right\}$, the average performance improvements are 4.14\%, 4.02\%, 4.44\%, and 5.17\%, respectively. These results verify the effectiveness of introducing recent features.

\begin{table*}[!t]
	\caption{The comparison results of old features and recent features on EPIC Kitchens.}
	\centering
	\resizebox{12cm}{!}{
		\begin{tabular}{ccccccccccccccc}
			\bottomrule[1pt]
			\multirow{3}*{$\Delta t$}&\multicolumn{7}{c}{Old features}&\multicolumn{7}{c}{Recent features}\\
			\cline{2-15}
			&\multicolumn{3}{c}{Top-5 Acc}&\multicolumn{3}{c}{ M.Top-5 Rec}&\multirow{2}*{Avg.}&\multicolumn{3}{c}{Top-5 Acc} & \multicolumn{3}{c}{ M.Top-5 Rec}&\multirow{2}*{Avg.}\\
             \cline{2-7}
			\cline{9-14}
			&VERB&NOUN&ACT&VERB&NOUN&ACT&&VERB&NOUN&ACT&VERB&NOUN&ACT&\\
			\cline{1-15}
      $\left\{0.4, 0.3, 0.2, 0.1 \right\}$&79.47&50.09&31.85&42.65&50.20&15.17&44.91&79.70&52.38&34.60&45.19&52.26&16.51&46.77\\
   $\left\{0.8, 0.6, 0.4, 0.2 \right\}$&79.33&49.77&31.99&43.26&50.50&15.24&45.02&79.82&52.38&34.90&45.28&52.49&16.08&46.83\\
   $\left\{1.2, 0.9, 0.6, 0.3 \right\}$&79.09&50.01&32.03&43.29&50.73&15.27&45.07&80.20&52.70&35.03&44.92&52.87&16.70&47.07\\
    $\left\{1.6, 1.2, 0.8, 0.4 \right\}$&79.11&49.85&32.72&43.45&50.06&15.04&45.04&80.26&53.10&35.45&44.84&53.43&17.14&47.37\\
			\bottomrule[1pt]
		\end{tabular}
	}

	\label{time}
	\vspace{-4mm}
\end{table*}
\begin{table}[!t]
\setlength\tabcolsep{3pt} 
	\caption{The results of the various duration of recent features on EGTEA Gaze+.}
	\centering
	\resizebox{6.5cm}{!}{
		\begin{tabular}{cccccccc}
			\bottomrule[1pt]
			\multirow{2}*{$\Delta t$}&\multicolumn{3}{c}{Top-5 Acc}&\multicolumn{3}{c}{ M.Top-5 Rec}&\multirow{2}*{Avg.}\\
             \cline{2-7}
			&VERB&NOUN&ACT&VERB&NOUN&ACT&\\
			\cline{1-8}
  	$\left\{0.4, 0.3, 0.2, 0.1 \right\}$&92.43&78.06&68.00&80.52&74.38&63.10&76.08\\
			$\left\{0.8, 0.6, 0.4, 0.2 \right\}$&92.25&77.78&67.13&80.33&73.32&62.25&75.51\\
			$\left\{1.2, 0.9, 0.6, 0.3 \right\}$&92.42&78.68&67.24&80.55&73.79&61.72&75.73\\
			\textbf {$\left\{1.6, 1.2, 0.8, 0.4 \right\}$}&93.11&79.20&69.03&83.16&75.22&64.65&77.40\\
			\bottomrule[1pt]
		\end{tabular}
	}

	\label{delta}
	\vspace{-1mm}
\end{table}
\noindent{\textit{\textbf{The effect of duration $\Delta t$.}}} Because $\Delta t$ plays a crucial role, we perform experiments with different settings of $\Delta t$. Tables~\ref{time} and \ref{delta} show that when recent features are extracted from segments corresponding to different $\Delta t$, the model's accuracy varies. In particular, when the $\Delta t$ values are $\left\{0.4, 0.3, 0.2, 0.1 \right\}$, $\left\{0.8, 0.6, 0.4, 0.2 \right\}$, $\left\{1.2, 0.9, 0.6, 0.3 \right\}$, and $\left\{1.6, 1.2, 0.8, 0.4 \right\}$, the average performances on EPIC are 46.77, 46.83, 47.07 and 47.37, and the average performances on EGTEA are 76.08, 75.51, 75.73, and 77.40.
In typical cases, when $\Delta t$ is larger, the performance is better because with a larger $\Delta t$, recent features involve more information.  
\begin{table}[t]
	\caption{The results of different Transformer configurations on EPIC-Kitchens.}
	\centering
	\resizebox{6.5cm}{!}{
		\begin{tabular}{ccccccc}
			\toprule[1pt]
			\multirow{2}*{Head/layer} & \multicolumn{3}{c}{Top-5 Acc}& \multicolumn{3}{c}{M.Top-5 Rec}\\
			\cline{2-7}
			&VERB&NOUN&ACT&VERB&NOUN&ACT\\
			\cline{1-7}
			2/1&80.08&\textbf{53.21}&35.43&44.26&52.67&16.42\\
			4/1&80.24&52.58&34.80&46.26&53.01&16.08\\
			6/1&80.12&52.40&34.66&\textbf{47.07}&53.10&15.91\\
			\midrule[0.5pt]
			\textbf{5/1}&\textbf{80.26}&53.10&\textbf{35.45}&44.84&\textbf{53.43}&\textbf{17.14}\\
			\midrule[0.5pt]
			5/2&79.94&52.62&34.30&45.69&53.37&15.97\\
			5/3&80.18&52.50&35.13&45.66&52.30&16.54\\
			\bottomrule[1pt]
		\end{tabular}
	}

	\label{transepic}
	\vspace{-3mm}	
\end{table}

\begin{table}[t]
	\caption{The results of different Transformer configurations on EGTEA Gaze+.}
	\centering
	\resizebox{6.5cm}{!}{
		\begin{tabular}{ccccccc}
			\toprule[1pt]
			\multirow{2}*{Head/layer}& \multicolumn{3}{c}{Top-5 Acc}& \multicolumn{3}{c}{M.Top-5 Rec}\\
			\cline{2-7}
			&VERB&NOUN&ACT&VERB&NOUN&ACT\\
			\cline{1-7}
			3/1&92.57&78.96&68.00&81.55&75.37&64.58\\
			4/1&\textbf{93.53}&78.93&68.01&82.20&75.06&63.99\\
	
			6/1&92.57&78.52&68.12&82.39&74.68&64.50\\

			\midrule[0.5pt]
			\textbf{2/1}&93.11&\textbf{79.20}&\textbf{69.03}&\textbf{83.16}&75.22&\textbf{64.65}\\
			\midrule[0.5pt]
			2/2&92.17&78.48&67.61&82.28&75.47&64.41\\
			2/3&92.73&78.90&67.71&81.19&\textbf{75.87}&63.85\\

			\bottomrule[1pt]
		\end{tabular}
	}
	\vspace{-1mm}	
	\label{transegtea}
\end{table}

\noindent{\textit{\textbf{The setting of Transformer.}}}
In our model, the Transformer enhances the complete-recent features, which are crucial for optimal performance. Our experiments, detailed in Table~\ref{transepic} for EPIC and Table~\ref{transegtea} for EGTEA, indicate the best performance with a Transformer head/layer configuration of 5/1 for EPIC and 2/1 for EGTEA. This setup accounts for the varying category numbers in these datasets, with more heads assigned to EPIC to manage its larger category diversity. 
Regarding the number of layer of the Transformer,  increasing the number of layers rapidly increases the parameters of the Transformer. As the Transformer parameters occupy a growing proportion of the overall model parameters, it might weaken the training for the GFL. Therefore, we found that setting the Transformer layer number to 1 achieves the optimal performance for our model.
\begin{table}[t]
	\caption{The results of multi-modal compositions on EPIC-Kitchens.}
	\centering
	\resizebox{6.8cm}{!}{
		\begin{tabular}{ccccccc}
			\bottomrule[1pt]
			\multirow{2}*{Modality}&\multicolumn{3}{c}{ Top-5 Acc} & \multicolumn{3}{c}{ M.Top-5 Rec} \\
			\cline{2-7}
			&VERB&NOUN&ACT&VERB&NOUN&ACT\\
			\cline{1-7}
			RGB&77.71&46.05&29.08&43.71&46.61&14.36\\
			FLOW&75.65&34.78&21.58&38.22&32.26&8.14\\
			OBJ&75.98&50.25&27.79&40.77&49.53&13.42\\
			RGB+FLOW&79.09&48.27&30.82&44.47&48.25&14.31\\
			RGB+OBJ&79.88&52.76&34.26&44.75&52.10&16.52\\	
			FLOW+OBJ&78.87&52.23&32.42&41.58&51.01&13.65\\
			\midrule[0.5pt]
			\textbf{RGB+FLOW+OBJ}&\textbf{80.26}&\textbf{53.10}&\textbf{35.45}&\textbf{44.84}&\textbf{53.43}&\textbf{17.14}\\
			\bottomrule[1pt]
		\end{tabular}
	}

		\vspace{-3mm}
	\label{co}
\end{table}

\noindent{\textit{\textbf{The effects of multi-modal compositions.}}}
Following several prior atudies \cite{furnari2019, sener2020, Zatsarynna2021MultiModalTC, 2021TransAction}, our model incorporates multiple modalities, including RGB (image feature), FLOW (optical flow feature), and OBJ (object feature) on the EPIC-Kitchens dataset. 
We assessed the impact of different modalities by conducting ablation experiments that considered individual modalities and their pairwise compositions. Table~\ref{co} presents the results. The table shows that the model performs best when all modalities are fused. By combining these modalities, our model can benefit from a more comprehensive understanding of the input.

\begin{figure*}[!h]
	\centering
	\setlength{\abovecaptionskip}{0.cm}
	\includegraphics[width=1.0\textwidth]{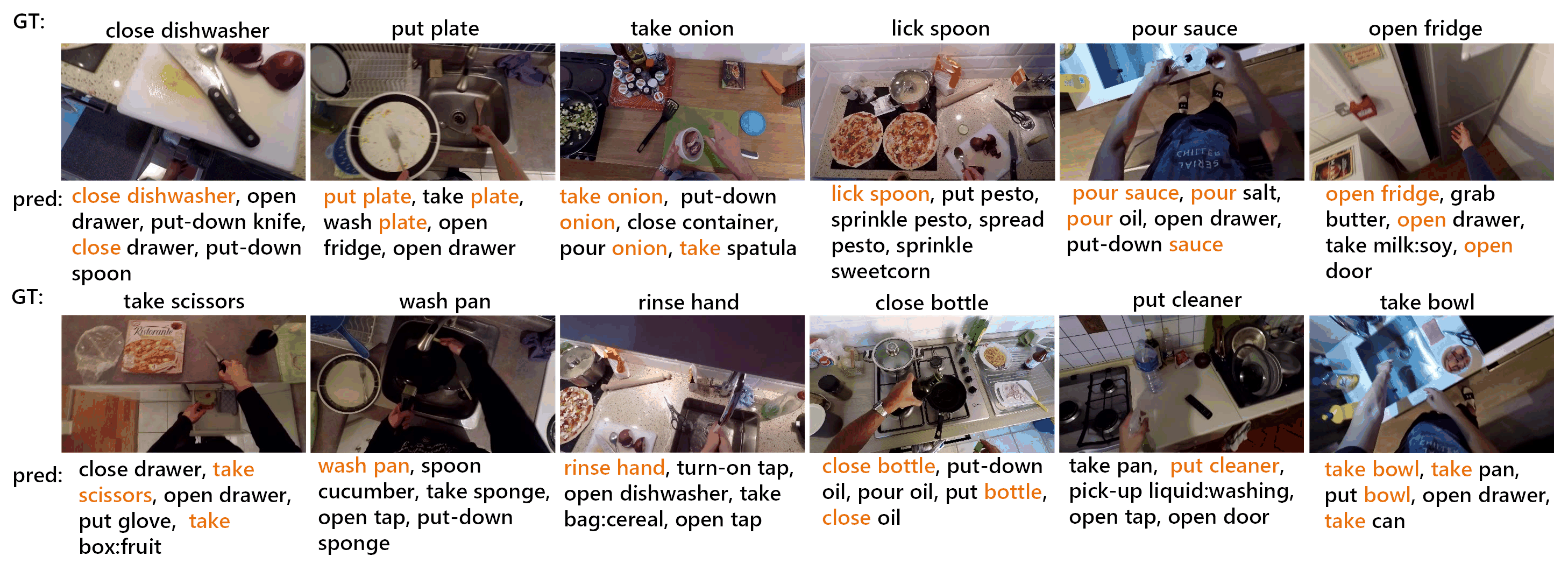}
	\caption{Prediction results of some examples. Orange font indicates the prediction that matches
the ground truth.}
	\label{fig_result}
\end{figure*}
\subsection{Qualitative results}
Our proposed method's intention prediction results (Fig.~\ref{fig_result}) highlight the model's effectiveness in challenging scenarios. Our model successfully generates accurate predictions despite the complexity of multiple objects and potentially misleading human behavior signals. For instance, in the close dishwasher scenario, the model correctly prioritizes the dishwasher over more visible objects, such as cutting boards and knives, aligning with the ground truth. Similarly, in the take scissors example, despite misdirection toward close drawer, the model still ranks the correct intention of take scissors highly. From the above observations, we can conclude that the HCR model demonstrates impressive performance, even in cluttered environments. This success can be attributed to its exceptional ability to leverage information from features spanning multiple temporal scales.

\section{Conclusion}
Herein, we study intention action prediction from an egocentric perspective. We propose an HCR information fusion model. In this model, we consider the unique information of $\boldsymbol{GGF}$s and local recent features. To fully use them, we propose a GFL mechanism to interactively update the recent and complete features through guidance, feedback, attention, residual, and other ideas. We use a Transformer to make the features obtain more information. The experimental results demonstrate the effectiveness of our model and provide several conclusions. 1) The GFL mechanism is crucial for models to strengthen feature representations, and 2) the complete–recent aggregation mechanism treats frames with different importance weights. Future research could explore multitask learning frameworks that jointly optimize intention prediction and complementary tasks, such as object detection, action segmentation, or action recognition. This study not only enhanced the robustness of the HCR model but also improved its generalizability across different scenarios.

\bibliographystyle{elsarticle-num-names} 
\bibliography{reference}

\end{document}